\documentclass[11pt,a4paper]{article}
\usepackage[hyperref]{naaclhlt2019}
\usepackage{times}
\usepackage{latexsym}
\usepackage{hyperref}
\usepackage{dirtytalk}
\usepackage{url}
\usepackage{graphicx} 
\usepackage{textcomp}
\usepackage{float}
\usepackage{subcaption}

\aclfinalcopy 

\title{Designing a Symbolic Intermediate Representation for Neural Surface Realization}

\author{Henry Elder \\
  ADAPT Centre \\
  Dublin City University \\
  {\tt henry.elder@adaptcentre.ie  } \\\And
  Jennifer Foster  \\
  ADAPT Centre \\
  Dublin City University \\
  {\tt  jennifer.foster@dcu.ie} \\\AND
  James Barry \\
  ADAPT Centre \\
  Dublin City University \\
  {\tt james.barry@adaptcentre.ie} \\\And
  Alexander O'Connor \\
  Autodesk, inc. \\
  {\tt alex.oconnor@autodesk.com} \\}

\date{}

\begin{document}
\maketitle
\begin{abstract}
Generated output from neural NLG systems often contain errors such as hallucination, repetition or contradiction.
This work focuses on designing a symbolic intermediate representation to be used in multi-stage neural generation with the intention of reducing the frequency of failed outputs.
We show that surface realization from this intermediate representation is of high quality and when the full system is applied to the E2E dataset it outperforms the winner of the E2E challenge.
Furthermore, by breaking out the surface realization step from typically end-to-end neural systems, we also provide a framework for non-neural content selection and planning systems to potentially take advantage of semi-supervised pretraining of neural surface realization models.
\end{abstract}

\section{Introduction}


For Natural Language Generation (NLG) systems to be useful in practice, they must generate utterances that are adequate, that is, the utterances need to include all relevant information.
Furthermore the information should be expressed correctly and fluently, as if written by a human.
The rule and template based systems which dominate commercial NLG systems are limited in their generation capabilities and require much human effort to create but are reliably adequate and known for widespread usage in areas such as financial journalism and business intelligence.
By contrast, neural NLG systems need only a well collected dataset to train their models and generate fluent sounding utterances but have notable problems, such as hallucination and a general lack of adequacy \citep{Wiseman2017ChallengesGeneration}.
There was a marked absence of neural NLG in any of the finalist systems in either the 2017 or 2018 Alexa Prize \citep{Fang2017SoundingSubmission,gunrock_alexa_prize_2018}.

Following prior work in the area of multi-stage neural NLG \citep{deep_syntax_trees_Ondrej_2016, mohit_navitagtion_2017,lapata_pipeline_aaai_2019,bender_mrs_generation_naacl_2019,plan_realization_yoav_NAACL2019}, and inspired by more traditional pipeline data-to-text generation  \citep{Reiter:2000:BNL:331955,gatt_kramer_nlg_journal_2018}, we present a system which splits apart the typically end-to-end data-driven neural model into separate utterance planning and surface realization models using a symbolic intermediate representation.
We focus in particular on surface realization and introduce a new symbolic intermediate representation which is based on an underspecified universal dependency tree \cite{underspecifiedUD_Mille_INLG2018}. In designing our intermediate representation, we are driven by the following constraints: 
\begin{enumerate}
\setlength\itemsep{-0.3em}
\item The intermediate representation must be suitable for processing with a neural system.
\item It must not make the surface realization task too difficult because we are interested in understanding the limitations of neural generation even under favorable conditions.
\item It must be possible to parse a sentence into this representation so that a surface realization training set can be easily augmented with additional in-domain data.
\end{enumerate}
%
%
%
%

%
%
Focusing on English and using the E2E dataset, we parse the reference sentences into our intermediate representation.
We then train a surface realization model to generate from this representation, comparing the resulting strings with the reference using both automatic and manual evaluation.
We find that the quality of the generated text is high, achieving a BLEU score of 82.47.
This increases to 83.38 when we augment the training data with sentences from the TripAdvisor corpus.
A manual error analysis shows that in only a very small proportion ($\sim$5\%) of the output sentences, the meaning of the reference is not fully recovered. This high level of adequacy is expected since the intermediate representations are generated directly from the reference sentences. An analysis of a sample of the adequate sentences shows that readability is on a par with the reference sentences.

Having established that surface realization from our intermediate representation achieves sufficiently high performance, we then test its efficacy as part of a pipeline system. On the E2E task, our system scores higher on automated results than the winner of the E2E challenge \citep{slug2slug_naacl_2018}. The use of additional training data in the surface realization stage results in further gains. These encouraging results  
suggest that pipelines can work well in the context of neural NLG.

\section{Methods}

\begin{figure}[ht]
    \centering
    \includegraphics[width=0.5\textwidth]{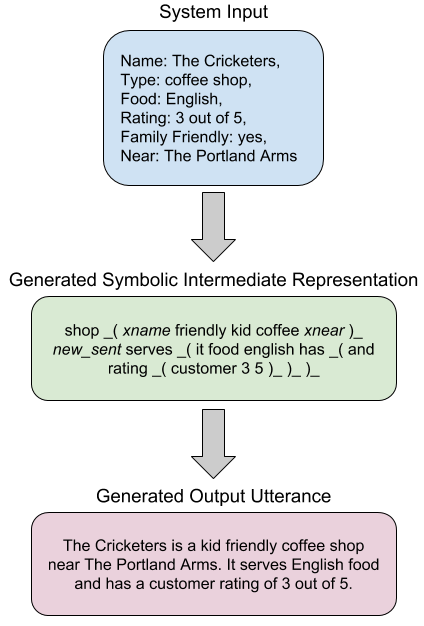}
    \caption{Example of two stage generation using the pipeline system. Both are real examples generated by their respective models.}
    \label{fig:pipeline}
\end{figure}
Our system consists of two distinct models.
The first is an utterance planning model which takes as input some structured data and generates an intermediate representation of an utterance containing one or more sentences.
The intermediate representation of each sentence in the utterance is then passed to a second surface realization model which generates the final natural language text.
See Figure \ref{fig:pipeline} for an example from the E2E dataset.
Both models are neural based.
We use a symbolic intermediate representation to pass information between the two models.

\begin{figure*}[ht!]
    \centering
    \includegraphics[width=1\textwidth]{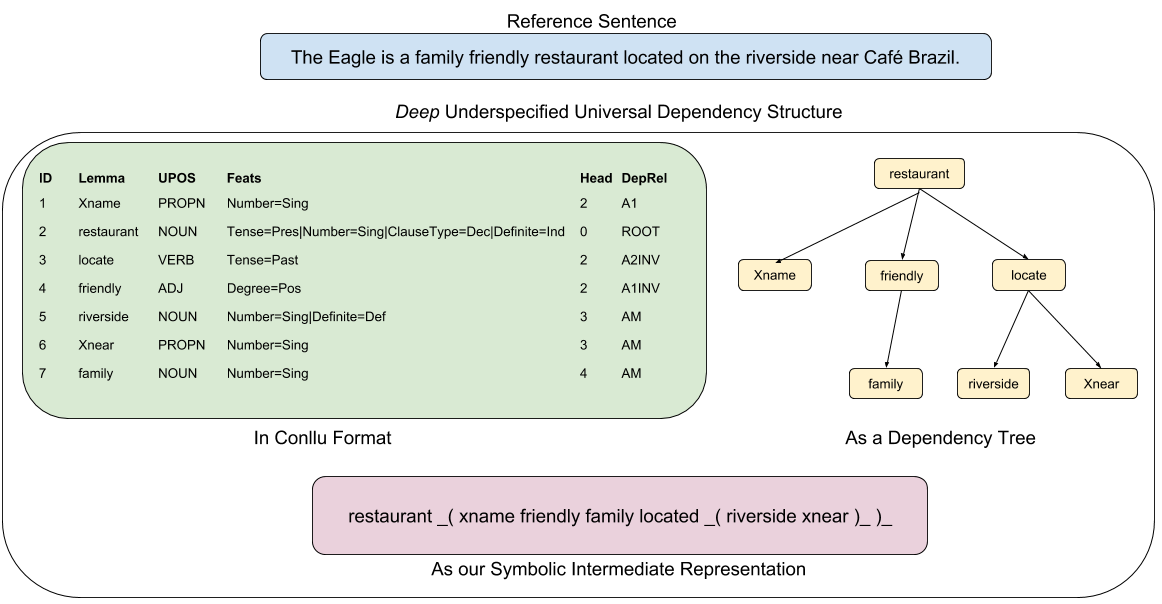}
    \caption{Different representations of the \emph{Deep} Underspecified Universal Dependency structure}
    \label{fig:representations}
\end{figure*}








\subsection{Symbolic Intermediate Representation}

The symbolic intermediate representation used is the \emph{deep}\footnote{\emph{deep} here is not referring to deep learning but rather as a contrast with another UUD variant known as the \emph{shallow} UUD. Shallow and deep surface realization tracks were used in both Surface Realization Shared Tasks \citep{mulit_surface_realization_2018,first_surface_realization_shared_task}} Underspecified Universal Dependency (UUD) structure \citep{underspecifiedUD_Mille_INLG2018}.
The UUD structure is a tree ``containing only content words linked by predicate-argument edges in the PropBank/NomBank \citep{propbank_2005,nombank_2004} fashion'' \citep{underspecifiedUD_Mille_INLG2018}.
Each UUD structure represents a single sentence.
The UUD structure was designed to ``approximate the kind of abstract meaning representations used in native NLG tasks'' \citep{underspecifiedUD_Mille_INLG2018}.
That is, the kind of output that a rule based system could be reasonably expected to generate as part of a pipeline NLG process.
However, to the best of our knowledge, no such system has yet been developed or adapted to generate the deep UUD structure as output.
Hence it was required to make a number of changes to the deep UUD structure during preprocessing to better suit a neural system designed to use the structure as a symbolic intermediate representation; namely we linearize the UUD tree, remove accompanying token features and use the surface form of each token, see Figure \ref{fig:representations}.
%





\paragraph{Linearization}
%
In order to use tree structures in a sequence-to-sequence model a linearization order for nodes in the tree must be determined.
Following \citet{neural_amr_konstas_ACL_2017} tree nodes are ordered using depth first search.
Scope markers are added before each child node.
When a node has only one child node we omit scope markers. Though this can lead to moderate ambiguity it greatly reduces the length of the sequence \citep{neural_amr_konstas_ACL_2017}.
%

When two nodes appear at the same level in the tree their linearization order is typically chosen at random, or using some rule based heuristic or even a secondary model \citep{thiago_surface_realization_2018}.
In this system linearization of equivalent level tokens is determined by the original order in which they appeared in the sentence.
We chose to use a consistent, as opposed to random, ordering of equivalent level nodes for the symbolic intermediate representation as it has been shown in a number of papers \cite{neural_amr_konstas_ACL_2017,slug2slug_naacl_2018} that neural models perform worse at given tasks when trained on symbolic intermediate representations sorted in random orders, even when that randomness is used to augment and increase the size of the data.
We chose to use original sentence order of tokens as the basis for ordering sibling nodes.
Though this is clearly a simplification, and gives the model additional information, it is an intuitive choice.
%

\paragraph{Features}
As well as the head id, tokens in the deep UUD structure are each associated with a number of additional features: dependency relations (DepRel), universal part-of-speech tag (UPOS) and lexical features (feats), see Figure \ref{fig:representations}.
Other neural based work on surface realization from the deep UUD structure included this information using factor methods \citep{Elder2018GeneratingModels}.
However our symbolic intermediate representation does not include these additional features.
By not including the additional features with each token we simplify the task of generating the symbolic intermediate representation using a neural model.
%
%
Token features could be generated using multitask learning as in \citet{pos_tag_prediction_2017} but we leave this for future work.
%

\paragraph{Lemmas vs. Forms}
In the deep UUD structure the token provided is a lemma, the root of the original form of a token.
Part-of-speech and lexical features are provided to enable a surface realization system to determine the form.
As we do not include these features in our symbolic intermediary representation we use the original form of token instead.
This is another simplification of the surface realization task.
While we found that \emph{lemma} + \emph{part of speech tag} + \emph{lexical features} typically provide enough information to reconstruct the original form, it is not a 100\% accurate mapping.

\section{Experiments}

\paragraph{Datasets}
Experiments were performed with the E2E dataset \citep{Novikova2017TheGeneration}.
Figure \ref{fig:pipeline} contains an example of the E2E input.
The E2E dataset contains a training set of 42,061 pairs of meaning representations and utterances.
Training data for the surface realization model was augmented, for some experiments, with the TripAdvisor corpus \citep{trip_advisor_corpus_2010}, which was filtered for sentences with a 100\% vocabulary overlap with the E2E corpus and a sentence length between 5 and 30 tokens, resulting in an additional 209,823 sentences, with an average sentence length of 10 tokens. By comparison the E2E corpus had sentence lengths ranging between 1 and 59 tokens with an average sentence length of 13 tokens. 

Both corpora were sentence tokenized and parsed by the Stanford NLP universal dependency parser \citep{qi2018universal}.
The parsed sentences in CoNLL-U format were then further processed by a special deep UUD parser \citep{underspecifiedUD_Mille_INLG2018}.
Utterances from the E2E corpus were delexicalised to anonymize restaurant names in both the \emph{name} and \emph{near} slots of the meaning representation.
All tokens were lower cased before training.

\paragraph{Models}
For the neural NLG pipeline system we train two separate encoder-decoder models using the neural machine translation framework OpenNMT \citep{opennmt_py_2017}.
We trained two separate encoder-decoder models for surface realization and content selection.
However both used the same hyperparameters.
A single layer LSTM \citep{Hochreiter1997} with RNN size 450 and word vector size 300 was used.
The models were trained using ADAM \citep{adam_learning_rate_2015} with a learning rate of 0.001.
The only difference between the two models was that the surface realization model was trained with a copy attention mechanism \citep{NIPS2015_5866}.

For the full E2E task a single planning model was trained on the E2E corpus.
However two different surface realization models were compared; one trained solely on sentences from the E2E corpus and another trained on a combined corpus of E2E and TripAdvisor sentences.
For baselines on the full E2E task we compare with two encoder-decoder models which both use semantic rerankers on their generated utterances; TGen \citep{deep_syntax_trees_Ondrej_2016} the baseline system for the E2E challenge and Slug2Slug \citep{slug2slug_naacl_2018} the winning system of the E2E challenge.

\paragraph{Automated Evaluation}
The E2E task is evaluated using an array of automated metrics\footnote{E2E NLG Challenge provides an official scoring script \url{https://github.com/tuetschek/e2e-metrics}}; BLEU \citep{Papineni:2002:BMA:1073083.1073135}, NIST \citep{nist_2002}, METEOR \citep{meteor_2007}, ROUGE \citep{Lin2004Rouge:Summaries}, and CIDEr \citep{Vedantam2015CIDEr:Evaluation}.
The two surface realization models were evaluated on how well they were able to realize sentences from the E2E validation set using silver parsed intermediate representations.
We report BLEU-4 scores\footnote{We input tokenized, lowercased and relexicalised sentences to the Moses multi-bleu perl script: \url{https://github.com/OpenNMT/OpenNMT-py/blob/master/tools/multi-bleu.perl}} for the silver parse generated texts from the surface realization models.
%
%
%
In both the E2E \citep{e2e_journal_paper2019} and WebNLG challenge \citep{automatic_metrics_correlation_design_Webnlg2018} it was found that automated results did not correlate with the human evaluation \emph{on the sentence level}.
However in the Surface Realization shared task correlation between BLEU score and human evaluation was noted to be highly significant \citep{mulit_surface_realization_2018}.

\paragraph{Manual Analysis}
The importance of using human evaluation to get a more accurate understanding of the quality of text generated by an NLG system cannot be overstated.
We perform human evaluation on the outputs of the surface realization model with a silver parse of the original utterances as input.
We evaluate the outputs first in terms of meaning similarity and then readability and fluency.
%
%

To evaluate the surface realization model we compare generated utterances with the human references.
For the meaning similarity human evaluation we remove sentences with no differences, only differences involving the presence or absence of hyphens or only capitalization differences. 
We evaluate meaning similarity between two utterances as whether they contain the same meaning.
We treat this a binary Yes / No decision as the generated utterances are using a silver parse and ought to be able to reconstruct a sentence that, while possibly differently structured, does express the same meaning.

We manually analyze failure cases where semantic similarity is not achieved to discover where the issues arise.
There may be failures in the method of obtaining the intermediate representation, in the surface realization model or some other issue with the intermediate representation.

We then pass on only those generated utterances deemed to have the same meaning with the reference utterance into the next stage of readability evaluation.
To evaluate readability we perform pairwise comparisons between generated utterances and reference utterances.
We randomize the order during evaluation so it is not clear what the origin of a particular utterance is.
We define readability, sometimes called fluency, as how well a given utterances reads, ``is it fluent English or does it have grammatical errors, awkward constructions, etc." \citep{mulit_surface_realization_2018}.
By investigating readability of utterances with meaning similarity, we hope to see how the surface realization model performs compared with a human written utterance.
The surface realization model is required to at least match human level performance in order to be usable, if it does not then we need to investigate where it fails and why.
We used Prodigy \citep{prodigy_tool_2018} as our data annotation tool.

\section{Results}

\subsection{Surface Realization Analysis}

\begin{table}[ht]
    \centering
    \begin{tabular}{ll}
    \hline
              & BLEU  \\ \hline
E2E           & 0.8247 \\
+ TripAdvisor & \textbf{0.8338} \\ \hline
    \end{tabular}
    \caption{Automated evaluation of surface realization models on validation set sentences}
    \label{tab:surface_realization_automated_results}
\end{table}

\paragraph{Automated evaluation}
To initially establish if training on additional data from a different corpus was beneficial we performed automated evaluation. 
Each surface realization model is provided a parse of the target sentence. 
The BLEU score is slightly higher, see Table \ref{tab:surface_realization_automated_results}, when the model is trained with the additional corpus data.

\paragraph{Manual analysis}

\begin{table}[ht]
    \centering
    \resizebox{0.5\textwidth}{!}{%
\begin{tabular}{lll}
\hline
                                          & E2E  & + TripAdvisor \\ \hline
Exact matches                             & 3807 & 3935          \\ \hline
Punctuation and/or determiner differences & 1242 & 1268          \\ \hline
\end{tabular}%
}
    \caption{Surface Realization of 8024 sentences in the E2E validation set}
    \label{tab:surface_realization_auto_manual_analysis}
\end{table}

\begin{table}[ht]
    \centering
    \resizebox{0.5\textwidth}{!}{%
    \begin{tabular}{lll}
\hline
                                  & E2E  & + TripAdvisor \\ \hline
Remaining sentences               & 2975 & 2821          \\ \hline
Sentences analysed                & 325  & 325           \\ \hline
Failed meaning similarities       & 76   & 45            \\
Same readability as reference     & 198  & 208           \\
Worse readability than reference  & 30   & 43             \\
Better readability than reference & 21   & 29            \\\hline
\end{tabular}%
}
    \caption{Manual analysis of a subset of remaining sentences from the 8024 sentences in the E2E validation set}
    \label{tab:manual_analysis_of_ms_and_read}
\end{table}


Starting with generated sentences from the E2E validation set, we first filter out exact or very close matches to the reference sentences, see Table \ref{tab:surface_realization_auto_manual_analysis}. Then taking a subset of remaining generated sentences, we establish that they contain the same meaning as the reference sentence. Finally we compare the readability / naturalness of the generated text with the human reference sentences, see Table \ref{tab:manual_analysis_of_ms_and_read}.

While the surface realization model trained on both E2E and Trip Advisor corpora generally outperforms the model trained on only E2E data, it has more sentences rated as \emph{Worse readability than reference}. More detailed manual analysis is required to tell whether this is a statistical anomaly or a true insight into how the additional data is affecting model performance.

\paragraph{Analysis of failed meaning similarities}

Looking at examples where a generated sentence  failed to correctly capture the meaning of the reference sentence we find the causes for this fall into a number of categories:

\begin{itemize}
    \item Poor sentence tokenization
    \item Problems with the reference sentence
    \item Unusually phrased reference sentence
    \item Unknown words
    \item Generation model failures (repetition or missing words)
\end{itemize}

The model trained on the additional TripAdvisor corpus has a larger vocabulary and has seen a wider range of sentences, and thus fails less often. Most failures appear to be due to reference sentences containing unknown tokens or being phrased in a new or unusual way the model has not seen before. 
A smaller number of cases are attributable to issues directly with the generation model, namely repetition or absence of tokens from the intermediate representation. Figure \ref{fig:fails} contains three examples of failed generation.

\begin{figure*}[th!]
    \begin{small}
    \centering
    \begin{subfigure}[b]{0.75\textwidth}
    \begin{itemize}
        \item[\textit{Ref:}] Do not go to The Punter near riverside.
        \item[\textit{IR:}] go \textunderscore( not xname riverside )\textunderscore
        \item[\textit{Gen:}] Not go to The Punter in riverside.
    \end{itemize}
    \caption{\centering Model generation failure}
    \hfill
    \label{fig:gen_fail}
    \end{subfigure}
    \begin{subfigure}[b]{0.75\textwidth}
    \begin{itemize}
        \item[\textit{Ref:}] With only an average customer rating, and it being a no for families, it doesn't have much going for it.
        \item[\textit{IR:}] have \textunderscore( rating \textunderscore( only average customer no \textunderscore( and it families )\textunderscore )\textunderscore it n't much \textunderscore( going it )\textunderscore )\textunderscore 
        \item[\textit{Gen:}] With a only average customer rating and its no families, it won't have much that going to it.
    \end{itemize}
    \caption{\centering Unusual phrasing in reference sentence} 
    \hfill
    \label{fig:unusual}
    \end{subfigure}
    \begin{subfigure}[b]{0.75\textwidth}
    \begin{itemize}
        \item[\textit{Ref:}] Have you heard of The Sorrento and The Wrestlers, they are the average friendly families.
        \item[\textit{IR:}] heard \textunderscore( you xnear \textunderscore( xname and )\textunderscore families \textunderscore( they average friendly )\textunderscore )\textunderscore
        \item[\textit{Gen:}] You can be heard near The Sorrento and The Wrestlers, they are average friendly families.
    \end{itemize}
    \caption{\centering Nonsensical reference sentence}
    \hfill
    \label{fig:nonsensical}
    \end{subfigure}
    \caption{Examples of reference sentences \textit{(Ref)}, intermediate representations \textit{(IR)} and generated texts \textit{(Gen)} from three different scenarios.}
    \label{fig:fails}
\end{small}
\end{figure*}

\begin{table}[ht]
\resizebox{0.5\textwidth}{!}{%
\begin{tabular}{llllll}
\hline
              & BLEU            & NIST            & METEOR          & ROUGE\_L        & CIDEr           \\ \hline
\multicolumn{6}{c}{Validation}                                                                          \\ \hline
TGen          & 0.6925          & 8.4781          & 0.4703          & 0.7257          & 2.3987          \\
Slug2Slug     & 0.6576          & 8.0761          & 0.4675          & 0.7029          & -               \\
Pipeline      & 0.7271          & 8.5680          & 0.4874          & 0.7546          & 2.5481          \\
+ TripAdvisor & \textbf{0.7298} & \textbf{8.5891} & \textbf{0.4875} & \textbf{0.7557} & \textbf{2.5507} \\ \hline
\multicolumn{6}{c}{Test}                                                                                \\ \hline
TGen          & 0.6593          & 8.6094          & 0.4483          & 0.6850          & 2.2338          \\
Slug2Slug     & 0.6619          & 8.6130          & 0.4454          & 0.6772          & 2.2615          \\
Pipeline      & 0.6705          & 8.6737          & \textbf{0.4573} & 0.7114          & 2.2940          \\
+ TripAdvisor & \textbf{0.6738} & \textbf{8.7277} & 0.4572          & \textbf{0.7152} & \textbf{2.2995} \\ \hline 
\end{tabular}%
}
\caption{Automated results on end-to-end task}
\label{tab:endtoend_automatic_results}
\end{table}

\subsection{End-to-End Analysis}

We report results on the full E2E task in Table \ref{tab:endtoend_automatic_results}. Both our systems outperform the E2E challenge winning system Slug2Slug \citep{slug2slug_naacl_2018}, with the system using the surface realization model trained with additional data performing slightly better. Both surface realization models received the same set of intermediate representations from the single utterance planning model.

Further human evaluation may be required to establish the meaningfulness of these higher automated results.

\section{Related Work}\label{Related Work}

The work most similar to ours is \citep{deep_syntax_trees_Ondrej_2016}.
It is also in the domain of task oriented dialogue and they apply two-stage generation; first generating deep syntax dependency trees using a neural model and then generating the final utterance using a non-neural surface realizer.
They found that while generation quality is initially higher from the two-stage model, when using a semantic reranker it is outperformed by an end-to-end seq2seq system.

Concurrent to this work is \citet{plan_realization_yoav_NAACL2019}.
In this work they split apart the task of planning and surface realization.
Conversely to \citet{deep_syntax_trees_Ondrej_2016} they employ a rule based utterance planner and a neural based surface realizer.
They applied this system to the WebNLG corpus \citep{webnlg_corpus_gardent_INLG2017} and found that, compared with a strong neural system, it performed roughly equally at surface realization but exceeded the neural system at adequately including information in the generated utterance.

Other work has looked for innovative ways to separate planning and surface realization from the end-to-end neural systems, most notably \citet{D18-1356} which learns template generation also on the E2E task, but does not yet match baseline performance, and \citet{negotiation_Liang_EMNLP2018} which has a dialogue manager control decision making and passes this information onto a secondary language generator.
Other work has attempted either multi-stage semi-unconstrained language generation, such as in the domain of story telling \citep{storygen_angela_fan_NAACL2019}, or filling-in-the-blanks style sentence reconstruction \citep{maskgan_goodfellow_ICLR2018}.






\section{Discussion}

Our system's automated results on the E2E task exceed that of the winning system.
This shows that splitting apart utterance planning and surface realization in a fully neural system may have potential benefit.
Our intuition is that by loosely separating the semantic and syntactic tasks of sentence planning and surface realization, our models are more easily able to learn alignments between source and target sequences in each distinct task than in a single model.
More clear alignments may help as the E2E corpus is a relatively small dataset, at least compared with dataset sizes used for neural machine translation \cite{bojar-EtAl:2018:WMT1} for which end-to-end neural models are the dominant paradigm.
Further human analysis of the generated utterances' fluency and adequacy\footnote{The generated utterance's coverage of the input meaning representation} could help determine what is driving the improved performance on automated metrics.

The design of our symbolic intermediate representation is such that additional training data can be easily collected for the surface realization model.
Indeed we see marginally better results on the E2E task with a surface realization model trained on both the E2E and TripAdvisor corpuses.
This approach could be further scaled beyond the relatively small number of additional sentences we automatically parsed from the TripAdvisor corpus.
In the E2E challenge it was noted that a semantic reranker was requisite for high performing neural systems \cite{e2e_journal_paper2019}.
Adding a semantic reranker to our system could likely help improve performance of the utterance planning step.

While we made simplifications to the intermediate representation, namely including forms over lemmas and using the original sentence order to sort adjacent nodes, their generation was still required to be performed by a higher level model.
It's possible that different higher level systems, for example a rule based utterance planning system, might prefer a more abstract intermediate representation.
Indeed this trade off between what information ought to go into the intermediate representation is a highly practical one.
A surface realization model trained using our automated representation could be made to work with a rule based system providing input.


\section{Conclusion}

We have designed a symbolic intermediate representation for use in a pipeline neural NLG system.
We found the surface realization from this representation to be of high quality, and that results improved further when trained on additional data.
When testing the full pipeline system automated results exceeded that of prior top performing neural systems, demonstrating the potential of breaking apart typically end-to-end neural systems into separate task-focused models.
\section*{Acknowledgements}
This research is supported by Science Foundation Ireland in the ADAPT Centre for Digital Con- tent Technology. The ADAPT Centre for Digi- tal Content Technology is funded under the SFI Research Centres Programme (Grant 13/RC/2106) and is co-funded under the European Regional De- velopment Fund.


\bibliography{refs}
\bibliographystyle{acl_natbib}

\appendix

\end{document}